\documentclass[]{article}
\pdfoutput=1
\usepackage{proceed2e}
\usepackage{times}
\usepackage{epsfig}
\usepackage{graphicx}
\usepackage{amsmath}
\usepackage{amssymb}
\usepackage{subfigure}

\newtheorem{mydef}{Definition}
\def\etal{\textit{et. al.}}

\title{Hashing Image Patches for Zooming}

\author{Mithun Das Gupta \\
        Epson Research and Development, Inc. \\
        2580 Orchard Parkway, Suite 225 \\
        San Jose, CA 95131. \\
        \emph{mdasgupta@erd.epson.com}} 

\begin{document}

\maketitle

\begin{abstract}
In this paper we present a Bayesian image zooming/super-resolution
algorithm based on a patch based representation.
We work on a patch based model with overlap and employ a Locally Linear Embedding
(LLE) based approach as our data fidelity term in the Bayesian inference. The image prior imposes continuity constraints across the overlapping patches. We apply an error back-projection technique, with an approximate cross bilateral filter. The problem of nearest neighbor search is handled by a variant of the locality sensitive hashing (LSH) scheme.
The novelty of our work lies in the speed up achieved by the hashing scheme and the
robustness and inherent modularity and parallel structure achieved by the LLE setup. The ill-posedness of the image reconstruction problem is handled by the introduction of regularization priors which encode the knowledge present in vast collections of natural images. We present comparative results for both run-time as well as visual image quality based measurements.
\end{abstract}

\section{Introduction}
Inferring a high resolution image from a single or a given set of
spatially coherent low resolution images has been a problem which
has been studied by numerous researchers in the past. Different
algorithms have been proposed to this end which find implementations
in the real world to handle different scenarios as well as end
goals. For instance, super-resolution of faces has an entirely
different set of literature which takes help from the vast field of
face identification and detection, whereas, super-resolution of
natural images still remains in the domains of interpolation and
optimization techniques.
Simple applications like increasing the
size of a digital image to complex scenarios like identifying
suspects from a surveillance video can all be modeled as some form
of super-resolution problem. Most of the techniques proposed till
now can be broadly classified into two main categories. Iterative
time/frequency domain methods and learning based techniques.
Iterative techniques can be traced back to the seminal work by
Tsai and Huang~\cite{tsai84}, which was then modified to the first
iterative implementation by Irani \etal~\cite{Irani91}. Iterative
methods use a Bayesian framework wherein an initial guess about
the high resolution frame is refined at each iteration. The image
prior is usually assumed to be a smoothness prior. Iterative
methods are inherently context independent and hence a few problems
encountered in cases like eyes in face enhancement still remain tough problems for the iterative methods. Geman and Geman~\cite{Geman84}, introduced an optimization
framework wherein a compatibility model of high-resolution
to low-resolution image features is constructed, and then the
inverse reconstruction problem can be solved by searching in the compatibility space.
Their work proposed modeling the unknown image as a Markov Random Field (MRF)
where neighboring image regions are assumed to be Markovian in
nature, thereby simplifying the inference problem. Given enough
training data, searching through compatibility space techniques have been
shown to work wonderfully well. Freeman et al~\cite{Freeman00}, studied natural
image super-resolution using Belief Propagation (BP)
algorithm~\cite{Pearl88} for inference in MRFs. The results clearly
show the superiority of the proposed method, but the run time for
the algorithm remained a bottleneck for the practical implementation
of their algorithm.

Another interesting approach was proposed by Chang
\etal~\cite{Chang04}, wherein they posed learning based
super-resolution as locally linear manifold estimation problem based
on the work by Roweis et al.~\cite{Roweis00}. Their method was based
on the fact that low-resolution image patches can be represented by
linear combinations of a few local patches which are the nearest
neighbors of the test patch in the low-resolution training set.
Corresponding high-resolution patches are then linearly
combined and the high-resolution patch is thereby inferred.
The convex set of coefficients learned for the low-resolution space
are translated as is to the high-resolution space. Boundary issues
were handled by simple averaging over pixel values. Example based
super-resolution methods have a similar flavor to the region growing based texture
synthesis technique proposed by Efros and Leung~\cite{Efros99}.
Texture synthesis can be assumed to be a sub-problem to generic super-resolution, which tries to exploit the regularities present in pure textures, wherein a candidate
region is {\it grown} based on sampling from the space of the nearest neighbors. This works efficiently for pure textures, but when the scene consists of greater details than one uniform texture, which is characterized by an inherent sense of repeatability, then the manifold based technique produces much better results. A similar method has been proposed for face enhancement, commonly known as hallucination, by Park \etal~\cite{Park07} wherein they propose linearity preserving projections. Manifold based techniques have also been used for successful image
retrieval as investigated by He \etal~\cite{He04}. But for most of
these methods the nearest-neighbor search still presents a
bottleneck for practical implementations.

In recent years, several researchers proposed to avoid the running
time bottleneck by using approximation
techniques~\cite{Arya94,Indyk98,Gionis99}. Locality Sensitive Hashing (LSH), which tries to address a similar problem was proposed by Indyk \etal~\cite{Indyk98}. The key idea is
to hash the points using several hash functions so as to ensure
that, for each function, the probability of collision is much higher
for objects which are close to each other than for those which are
far apart. One can determine near neighbors by hashing the
query point and retrieving elements stored in buckets containing
that point. The algorithm proposed in~\cite{Gionis99} worked in the
{\it hamming space} and such it assumed the data points to be in $\{0,1\}^d$ space.
In this work we extend the work of Datar \etal~\cite{Datar04}. Their work introduced
the idea of {\it p-stable} distributions to generate the hash
functions and showed that this technique can be implemented directly
on the data space without transformation to the hamming space.

\section{Motivation}
In this paper we present a method which builds on the algorithm
proposed by Chang~\etal~\cite{Chang04}, with a few key differences
which render this work unique and the first of its kind to the best
of our knowledge.
The nearest neighbor estimation step mentioned in the
original work by Roweis \etal~\cite{Roweis00}, has been
used as is in the work by Chang \etal~\cite{Chang04}. This technique
presents the most crucial bottleneck in the scalability as well
practical use of their algorithm. We counter this deficiency by employing Locality
Sensitive Hashing scheme which works well in scenarios where
nearest neighbor (NN) problem can be approximated by an
$(R;c)$-NN problem~\cite{Datar04}, defined later in Sec~\ref{Sec:LSH}. For the method proposed by Chang \etal~\cite{Chang04}, the image reconstruction quality quickly deteriorates due to the lack of effective regularization. We counter this by a back-propagation technique, where we introduce an optimized approximation of the bilateral filter which has been shown to preserve edges while simultaneously suppressing noise.


The rest of the paper is organized as follows. We introduce the
notations in Sec~\ref{SSec:Notation} which would be consistently
used for the rest of this paper. In Sec~\ref{Sec:LLE}, Sec~\ref{Sec:LSH} and Sec~\ref{Sec:BLF} we introduce the mathematical details for the three
main techniques which form the crux of our work.
Sec~\ref{Sec:Method}, illustrates the main body of this work. We enlist the modifications and enhancements made to each of the techniques mentioned earlier, to make them work for our application. We present the experimental details in
Sec~\ref{Sec:Expts}. 
Finally, we conclude our work with
discussions about ongoing and future work in Sec~\ref{Sec:Future}.

\subsection{Notations}\label{SSec:Notation}Let
$\{X_1,X_2\dots,X_T\}\in\mathbb{\mathbf{X}}$ denote the set of
high-resolution images, from which we generate the corresponding set
of low-resolution images
$\{Y_1,Y_2\dots,Y_T\}\in\mathbb{\mathbf{Y}}$ by first blurring by a
Gaussian kernel $\mathcal{K}(\sigma^2)$, where $\sigma^2$ denotes
the spread of the kernel, and then down-sampling by a factor
$\mathcal{D} = 2$, when not specified explicitly. The low-resolution
images are then interpolated back to the original size of the
high-resolution images, so that corresponding images are of the same
size. Individual low and high resolution image pairs are then
divided into overlapping patches denoted by $x_i^j$, $y_i^j$, where
$i\in\{1,2,\dots,T\}$ denotes the image index, and
$j\in\{1,2,\dots,N_p\}$ denotes the patches in the $i^{th}$ low/high
image pair in the training data. We will drop the image index in the
later sections when it is clear that only patch index is required to
identify the low/high resolution patch pairs. The patches are
numbered in row-principle order for each image. Note, that the patch
index runs through all the images in the training set assigning
\textbf{\textit{unique integer ID's}} to each patch from each image
in the training data. The notation $N_p$ specifically denotes the
total number of patches generated for the entire data set. Also note
that there are $N_p$ high resolution patches and $N_p$
low-resolution patches which need to be explicitly stored onto the
disk.

\section{Background}
\subsection{Locally Linear Embedding (LLE)}\label{Sec:LLE}LLE
as a manifold learning technique was introduced by Roweis
\etal~\cite{Roweis00}, wherein they proposed a local neighborhood
based linear estimation technique to unfold complex manifolds. The
key idea is to compute locally linear weights to approximate the
actual high-dimensional space. This method has found many
applications in the fields of machine learning and computer vision,
and Chang \etal~\cite{Chang04} were one of the first group of
researchers to propose an LLE based algorithm for image
super-resolution.

Suppose there are $N$ points in a high-dimensional space of
dimension $D$. The $N$ points are assumed to lie on a manifold of
dimension $d$, where $d \ll D$. The principle assumption of this
technique is that if the manifold is {\it smooth}, meaning the local changes in the topology can be assumed to be linear, then a sample on the manifold can be approximated by a linear combination of a few of its neighbors. As such the two main components of the LLE algorithm are
\begin{enumerate}
\item[1.]For each data point in the D-dimensional data
space:
\begin{enumerate}
\item
Find the set of $K$ nearest neighbors in the
same space.
\item
Compute the reconstruction weights of the
neighbors that minimize the reconstruction
error.
\end{enumerate}
\item[2.]Compute the low-dimensional embedding in the $d$ dimensional
embedding space such that it best preserves
the local geometry represented by the reconstruction
weights.
\end{enumerate}

\subsection{Locality Sensitive Hashing (LSH)}\label{Sec:LSH}
Nearest neighbor problem can be relaxed by posing it as an $(R;
c)$-NN problem, where the aim is to return a point $p$ which is at a
distance $cR$ from the query point $q$, when there is a point in the
data set $P$ within distance $R$ from $q$. LSH technique was
introduced by Indyk et. al.~\cite{Indyk98}, to solve the $(R; c)$-NN problem .
For a domain $S$ of the points set with distance measure $D$, an LSH
family is defined as:
\begin{mydef}
A family $\mathbb{H} = \{h : S \rightarrow U\}$ is called $(r_1; r_2; p_1; p_2)-$
sensitive for $D$ if for any $v$, $q \in S$
\begin{eqnarray*}
  if v \in B(q; r_1) &then& Pr_\mathbb{H}[h(q) = h(v)] \ge p_1 \\
  if v \notin B(q; r_2) &then& Pr_\mathbb{H}[h(q) = h(v)] \le p_2.
\end{eqnarray*}
where $B(q; r_1)$ is the ball of radius $r_1$
centered at $q$ and $p_1>p_2$.
\end{mydef}
$k$ such hash functions are concatenated to form a new function
class $\mathbb{G} = \{g : S \rightarrow U^k\}$ such that $g(v) =
(h_1(v), h_2(v),\dots,h_k(v))$, where $h_i \in \mathbb{H}$. $L$ such
functions $(g_1,g_2,\dots,g_L)$ are chosen independently uniformly
at random and form the buckets into which the training patches are
hashed. During testing the test patches are hashed into the buckets
and all the matching training patches are extracted as the nearest
neighbors.

\subsection{Bilateral filter}\label{Sec:BLF}
Bilateral filtering~\cite{Tomasi98}, is a non-linear filtering technique which
combines image information from spatial domain
as well as the feature domain. It can be
represented by the following equation
\begin{equation}\label{EQN:bl_filter}
\mathbf{h}(x)=\frac{1}{k(x)}\sum_{y\in \Gamma(x)}\mathbf{I}(y)c(x,y)s(\mathbf{I}(x),\mathbf{I}(y))
\end{equation}
where $\mathbf{I}$ and $\mathbf{h}$ are the input and the output images respectively, $x$ and $y$ are pixel locations, and $\Gamma(x)$ is the neighborhood around the pixel $x$.
\begin{equation}\label{EQN:normalization}
    k(x)=\sum_{y\in \Gamma(x)}c(x,y)s(\mathbf{I}(x),\mathbf{I}(y))
\end{equation}
is a normalization term at pixel $x$. Normally the function $c(.)$ is defined as
\begin{equation}\label{EQN:cx}
    c(x,y)=\exp\frac{-\|x-y\|_2^2}{2\sigma_c^2}
\end{equation}
and similarly the filter $s(.)$ is defined as
\begin{equation}\label{EQN:sx}
    s(x,y)=\exp\frac{-\|x-y\|_2^2}{2\sigma_s^2}
\end{equation}

\section{Bayesian Inference with back-propagation}\label{Sec:Method}
Denote the input low resolution (LR) image as $I_L$ and the true high resolution (HR) image as $I_H$. Now we can write
\begin{equation}
p(I_H|I_L) \propto p(I_L|I_H)p(I_H)
\end{equation}
The maximum likelihood estimate then gives an estimate for the best $I_H$. the error image which is the deviation from the current inference to the observed image is then computed as
\begin{equation}\label{EQN:Err_img}
e_r(I) = (I_H\star g)\downarrow -~ I_L
\end{equation}
Here $g$ is a smoothing kernel and $\star$ represents the convolution in spatial domain.
Back-propagation technique, introduced by Irani \etal~\cite{Irani91}, adds a smoothed version of the error image to the current estimate of the HR image.
\begin{equation}
    I_H^{t+1} = I_H^{t} + e_r(I)\uparrow \star ~p
\end{equation}
$p$ is the back-projection filter and can be similar to $g$. We substitute a bilateral filter kernel for the back-projection filter as mentioned in the forthcoming sections.

For a patch based representation, assuming Gaussian noise model, the negative log likelihood term can be written as
\begin{equation}\label{EQN:Dist}
 -\log p(I_L|I_H) = \sum_{n = 1}^ND(I_L^n,I_H^n)
\end{equation}
where $N$ is the total number of patches of size $p\times p$ in the image and $D(y,x)$ is a metric discussed in detail in the later sections. We also assume that the sample rate discrepancies have been absorbed into $D(.)$. The image prior is
assumed to be a smoothness constraint imposed on the overlapping boundary of the patches.
For two neighboring patches in the HR image the image prior is denoted as
\begin{equation}
 \log p(I_H) = \sum_{neighbors(p,q)}\frac{-\|I_H^p-I_H^q\|^2_{overlap}}{2\sigma_o^2}
\end{equation}
Now let us look at the likelihood term in greater detail. For each patch in the LR image, we find the closest patches in the
LR dataset and compute the LLE reconstruction weights. Then, we impose the same reconstruction weights on the
corresponding HR patches and reconstruct the HR patch. The search in the LR space is usually the speed bottleneck for such methods. We implement a variant of the LSH technique to alleviate this problem.


\subsection{Hashing}
Datar \etal~\cite{Datar04}, showed that Gaussian distribution is a
\textit{p-stable} distribution under $l_2$ norm, and hence preserved
distances after projection by a Gaussian random vector. It has been
proved that projection by Cauchy random vectors preserve $l_1$
distances~\cite{Zolotarev86}. Recent research work on
the statistics of natural images~\cite{Olshausen96,Weiss01}, has shown that natural images have sparse derivatives, which follow an approximate Laplacian distribution. This hints at an $l_1$ type of distance metric for the nearest neighbor searching. Based on this knowledge we chose the following formulation for our hash function
\begin{equation}\label{Eqn:hash_fn}
    h_{\boldsymbol{\alpha},\beta,r}(\mathbf{v}) = \lfloor \frac{\boldsymbol{\alpha}.\mathbf{v} + \beta}{r} \rfloor
\end{equation}
where $r$ is an integer chosen
to balance the hash bins, $\beta$ is a uniform random number chosen
from $[0,r]$ and $\boldsymbol{\alpha}$ is a Cauchy random vector whose elements
are sampled from the rule
\begin{eqnarray}\label{Eqn:cauchy}
	\nonumber
	a &\sim& \mathcal{N}(0,1) \\
	\nonumber
	b &\sim& \mathcal{N}(0,1) ~s.t. \|b\| \ge \epsilon \\
	\alpha_i &=& a / b
\end{eqnarray}
where $\mathcal{N}(0,1)$ represents a normal distribution. $\mathbf{v}$ is some feature representation of a low-resolution patch $y_i^j$ and $\lfloor x \rfloor$ denotes the integer less than or equal to $x$. We concatenate $k=3$ such hash
functions and form $1$ hash table. Based on the $k$ integers we form
the bins in the table and store the integer ID corresponding to the
patch $y_i^j$ into the hash table. $L=30$ such hash tables are
formed for matching the test patches against the training patches.
We introduce a \textit{hash match} criterion during matching which
states that a matching patch ID found in one hash table should be
present in at least $t$ out of $L$ hash tables.

{\bf{Feature representation:}} The low-resolution patches
are represented as gradient based features for hashing. We use
different feature representations ranging from first order gradient
(convolution with [$-1, 0, 1$], $[-1, 0, 1]^T$) to second
order gradient (convolution with [$-1, 0, 2, 0, -1$], $[-1, 0, 2, 0, -1]^T$).

{\bf{Hashing Quality (determining $\mathbf{r}$):}} Entropy has been
one of the most widely accepted metrics for measuring the quality of
hashing. In our case entropy directly translates to the speed of
generating the list of all the approximate nearest neighbors. But
improving the entropy can sometimes lead to poor image
reconstruction quality, which hints at a tradeoff for practical
implementations. The denominator $r$ in Eqn~\ref{Eqn:hash_fn}, is
used to control the quality of hashing,
\subsection{Locally Linear Embedding}Let us denote the test patch as $y^t$.
We query for the test patch into each hash table and collect all the
$y^m$ matches found, where $m\in {1,2,\dots,M}$. We purposefully
drop the image index as introduced in Sec~\ref{SSec:Notation},
because after the training stage all patches are treated equally,
independent of their originating images. If $M$ is larger than some
pre-defined number of matches (typically $3\times L$) further
searching is stopped. Once the matches have been identified, we
solve the following constrained least square minimization problem
\begin{eqnarray}\label{Eqn:LinSys}
  \min_{W} && \| y^t - \sum_{j=1}^Mw_my^m \|^2 \\
  st && \sum w_m = 1
\end{eqnarray}
Once, the weight matrix is computed the high-resolution patch
$x^t$ corresponding to $y^t$ can simply be written as
\begin{equation}\label{Eqn:HR_Rec}
    x^t = \sum_{j=1}^Mw_mx^m
\end{equation}
For, all the results shown later in the paper, we have used $5\times
5$ patches with one pixel width overlap. 
Going back to the metric $D(.)$ introduced in Eqn.~\ref{EQN:Dist}, the metric is equivalent to solving the LLE reconstruction equation in Eqn.~\ref{Eqn:LinSys}, based on the neighbors found by the hashing scheme.
\subsection{Approximate Bilateral Error Propagation}
Normally, the filter $s(.)$ is defined as in Eqn.~\ref{EQN:sx}. But in this paper, we present a different formulation for this filter. The main concern with numerous
researchers with the simple form of bilateral filter is the fact that, the complexity of
the filter grows as $O(r^2)$. 
In view of the kernel size bottleneck, we define the filter $s(.)$ as
\begin{equation}\label{EQN:sx2}
    s(\mathbf{I}(x),\Gamma(x))=\exp\frac{-\|\mathbf{I}(x)-\frac{1}{[\Gamma(x)]}\sum_{{y\in \Gamma(x)}}\mathbf{I}(y)\|_2^2}{2\sigma_s^2}
\end{equation}
where the notation $[.]$ just denotes the cardinality of the neighborhood
We also borrow from Paris et. al.~\cite{Paris09}, and adopt the idea of cross bilateral filtering wherein, $\mathbf{I}(x)$ and $\mathbf{I}(y)$ can come from different images. Since this filter operates on the error image, we replace $\mathbf{I}(y)$ terms with $\mathbf{\hat{I}}_e(y)$, where $\mathbf{\hat{I}}_e$ represents the error between the original LR image and bi-cubic interpolated HR image. The new formulation of the range filter $s(x)$ can now be exploited, since the mean over the filter kernel, can be computed extremely efficiently. Again, borrowing from Porikli~\cite{Porikli08}, we adopt the integral image technique, which renders the evaluation of the filter $s(.)$ in Eqn.~\ref{EQN:sx}, an $O(1)$ operation. The new filter can now be written as
\begin{equation}\label{EQN:bl_filter_2}
\mathbf{h}(x)=s(\mathbf{I}(x),\Gamma(x))\frac{\sum_{y\in \Gamma(x)}\mathbf{I}(y)c(x,y)}{k(x)}
\end{equation}
Assuming a normalized filter $C$, and denoting the filter response for $s(.)$ for the entire image as $\mathbf{I}_s$, we can write the filter response succinctly as
\begin{equation}\label{EQN:bl_filter_3}
\mathbf{H}=\mathbf{I}_s \otimes (\mathbf{I} \star C)
\end{equation}
where $\otimes$ means an elementwise multiplication.


\section{Experiments and Results}\label{Sec:Expts}
\subsection{Training Data Generation}\label{SSEC:Tr_Data}
\begin{figure}[hbt]
  \begin{center}
    \includegraphics[width=0.45\textwidth]{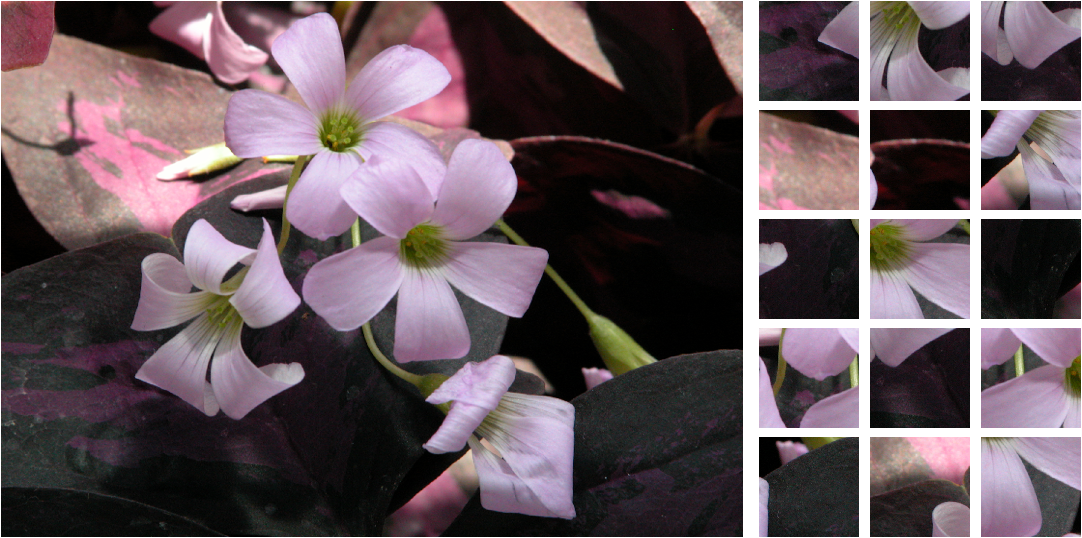}
  \end{center}
  \caption{Generation of training data. Left: original image, right: gradient based sampling of smaller sub-images.}
  \label{fig:tr_im}
\end{figure}

One of the key benefits of our method over others is the fact that it can handle relatively large amounts of training data. Because of the approximate nearest neighbor search, facilitated by the LSH formulation, the inference algorithm only looks at a sub-part of the training data. To generate the training data for the experiments shown in Fig~\ref{fig:dog_brown}, we collected $500$ high resolution images from~\cite{McGill04}.
We compute the horizontal and vertical first order gradient at each pixel location of an image and then generated a cumulative distribution function based on the gradient values. We sampled for the center pixel location and then saved  a $100 \times 100$ image centered at this pixel. $25$ such samples were drawn at random from each of the $500$ images chosen form~\cite{McGill04} and then saved as training images. Each training image is then blurred and downsampled to form the low-resolution counterpart for the high resolution training images. For a patch size of $5 \times 5$, with an overlap of 2 pixels, we collect a total of around $800$ patch pairs from one image. The total number of training patch pairs is of the order of $800 \times 25 \times 500\sim 10^6$ patch pairs. Examples of the sub-images collected from one image is shown in Fig~\ref{fig:tr_im}. For all the results none of the test images were present in the training set. For Fig~\ref{fig:freeman}, we added a small part of the image in the training set.
\subsection{Comparative Results}
We present the comparative results for our algorithm against standard image interpolation techniques like bicubic interpolation.
We also compare our results against similar learning based techniques like Chang \etal~\cite{Chang04} and Freeman's
method~\cite{Freeman00}. We present comparative run-time analysis
against Chang~\cite{Chang04}, where we count one nearest
neighbor as one operation. Our analysis clearly proves the time
efficiency achieved by incorporating the hashing scheme into the
linear embedding technique used in~\cite{Chang04}.

\begin{figure}
     \centering
      \subfigure[low resolution]{
          \includegraphics[width=0.2\textwidth]{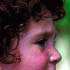}}
      \subfigure[high resolution]{
          \includegraphics[width=0.2\textwidth]{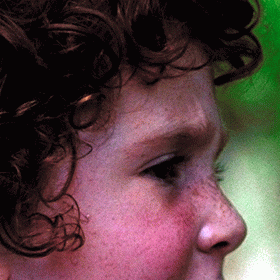}}\\
      \subfigure[Freeman's\cite{Freeman00} method]{
          \includegraphics[width=0.2\textwidth]{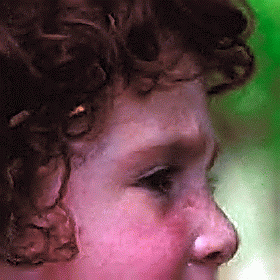}\label{fig:fr_res}}
      \subfigure[our method]{
          \includegraphics[width=0.2\textwidth]{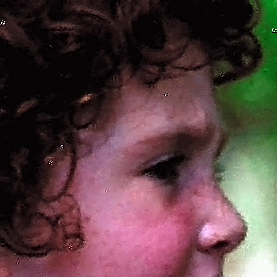}}
      \caption{4X magnification for the girl image.}
     \label{fig:freeman}
\end{figure}

The first set of results were performed on the little girl image which was presented in~\cite{Freeman00} and has since been used by numerous researchers for comparison. Fig~\ref{fig:freeman} presents the comparative results of our method against~\cite{Freeman00}. Note that Fig~\ref{fig:fr_res} has been taken from the author's website\footnote{http://www.ai.mit.edu/people/wtf/superres}. One primary benefit of our method over~\cite{Freeman00} is the fact that our method goes through all the patches in the test image just once, compared to an iterative belief propagation routine in~\cite{Freeman00}, and hence can handle large amount of training data, while still being able to generate acceptable results. 40,000 training patch pairs were reported in~\cite{Freeman00} and they run the process for upto 10 iterations. The run time for their method on our system (Pentium 4, 3Mhz, 2MB RAM) was of the order of 6 minutes for one iteration, which results to around 60 minutes runtime for an image which is of size $280 \times 280$. Our method runs in one iteration and takes around 1 minute for generating the results. Note that our training set for this experiment contained around 300,000 patch pairs, which is roughly 3 times that handled by~\cite{Freeman00}.

\begin{figure}[hbt]
  \begin{center}
    \includegraphics[width=0.45\textwidth]{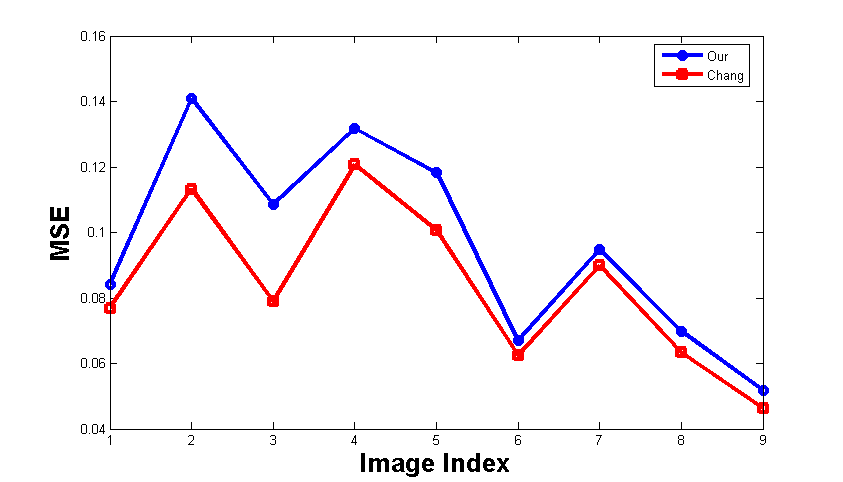} \\ 
    \includegraphics[width=0.45\textwidth]{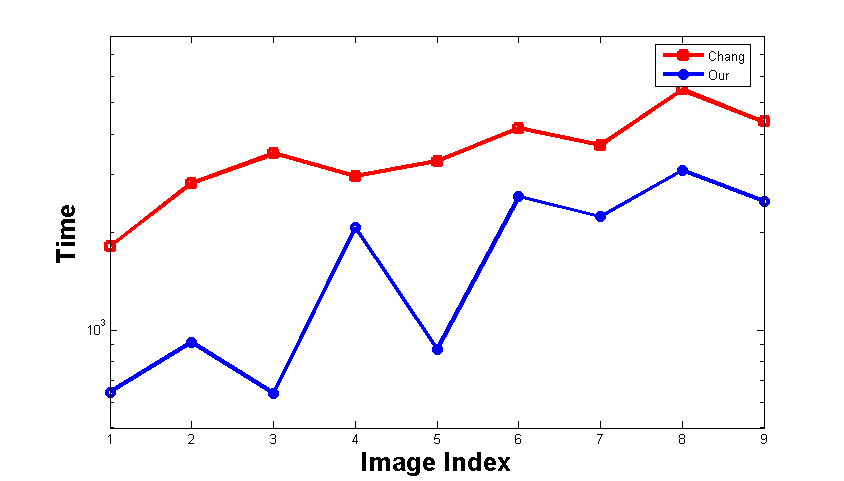}
\end{center}
  \caption{Comparison of our method against Chang \etal~\cite{Chang04}, Top: Mean square error, Bottom: run time in log scale. Note that the best result (image 3) points to a time saving by a factor of 100.}
  \label{fig:comp_chang}
\end{figure}

We compare our run times against Chang \etal~\cite{Chang04} as presented in Fig~\ref{fig:comp_chang}. As hinted earlier, we denote each similarity against a patch in the training data base as one time unit and count the number of similarity checks we need to perform to generate the results. Since comparing against the training patches is the most important part of the algorithm (and also the most time consuming one), this can be treated as a fair comparison for the run times of the two algorithms.
We compute mean number of similar patches generated by our algorithm to infer all the patches of the test image. For~\cite{Chang04} this number remains fixed at the total number of patches in the training set.
Fig~\ref{fig:comp_chang} (top) shows the MSE for $9$ different experiments averaged $20$ times each. The two curves are almost similar strengthening our claim that the approximate nearest neighbor hashing scheme performs almost at par with the complete nearest neighbor scheme. On average we underperform by a factor of about $0.025$ (for images which are all larger than $200 \times 200$ pixels)which is negligible compared to the time gain which we accomplish by our method. Fig~\ref{fig:comp_chang} (bottom) shows the comparison of the quality factor $Q$ for the two methods. Note that the quality comparison results are shown in log scale to reduce the wide gain achieved by our method over Chang's~\cite{Chang04} method, and plot both the curves in the same figure. As mentioned earlier, we learn $5$ hash tables and hence deal with $5$ times the training data, but still manage to generate improvements in run time by orders of magnitude. The maximum improvement in runtime is around 2 on the log scale which is about 100 times. The average time gain is about 70 times over Chang's~\cite{Chang04} method. The run times reported in this paper are all generated on a single machine. The structure of the algorithm can be exploited to break the problem into similar sub-problems and can be run on different machines.

\begin{figure}
  \begin{center}
   \includegraphics[width=0.5\textwidth]{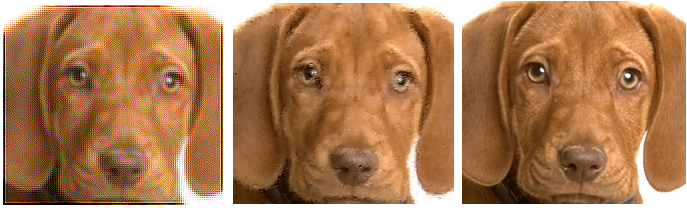}\\
   \includegraphics[width=0.5\textwidth]{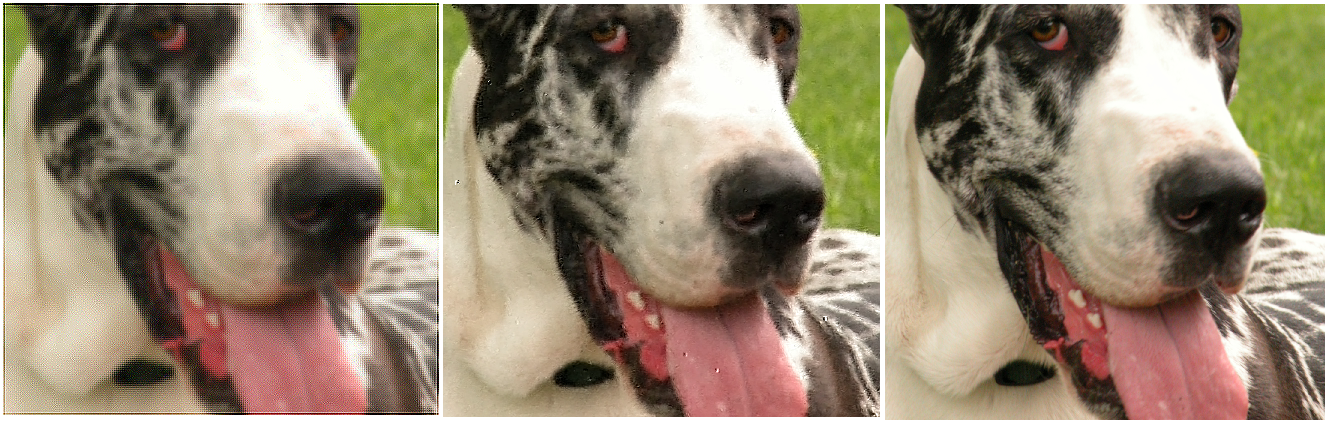}\\
    \includegraphics[width=0.5\textwidth]{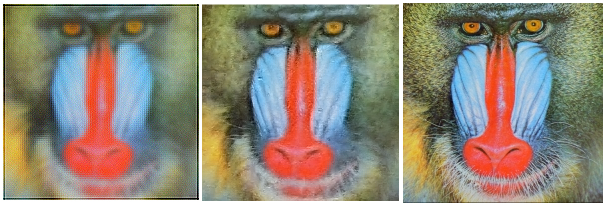}\\
    \includegraphics[width=0.5\textwidth]{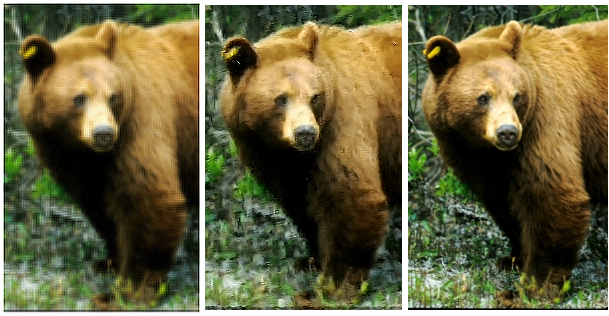}
  \end{center}
  \caption{Left: bicubic+unsharp, centre: our result, right: original image.}
  \label{fig:dog_brown}
\end{figure}

\begin{figure}
  \begin{center}
    \includegraphics[width=0.5\textwidth]{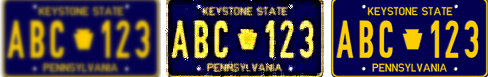}
  \end{center}
  \caption{De-blurring experiment. Left: input image. center: our result, right: original image.}
  \label{fig:denoise}
\end{figure}

\begin{figure*}
  \begin{center}
    \includegraphics[width=1.0\textwidth]{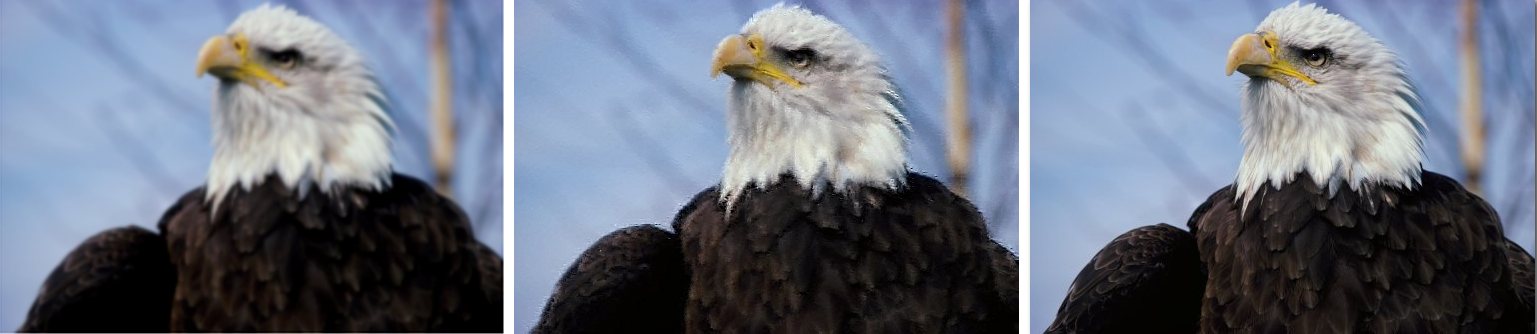}
  \end{center}
  \caption{Left: input image. center: our result, right: original image.}
  \label{fig:short}
\end{figure*}

Next set of results compare our method against bicubic interpolation as shown in Fig~\ref{fig:dog_brown}. Since, 
bicubic interpolation by itself cannot match our results, we perform an additional {\it{unsharp}} operation and report the comparative results in Fig~\ref{fig:dog_brown}. We also perform similar experiments for de-blurring and zooming. Fig~\ref{fig:denoise} shows input blurred image and the result of our algorithm along with the original HR image. Fig~\ref{fig:short} shows the input image stretched two times and the result of our algorithm along with the original HR image.


\section{Conclusion and Future work}\label{Sec:Future}
In this paper we present a truly large scale system able to handle large amounts of data in acceptable time for image zoomiong/super-resolution. We introduce the idea of LSH for feature based image patch hashing for efficient approximate nearest neighbor identification in close to constant time. An LLE based scheme, is then employed to infer the high-resolution patches. We present image reconstruction results as well as run-time comparisons against established learning based methods.

{\small
\bibliographystyle{ieee}
\bibliography{egbib}
}

\end{document}